\documentclass{article}
\usepackage[utf8]{inputenc}
\usepackage{amsmath, amssymb}
\usepackage{graphicx}
\usepackage{hyperref}
\usepackage{booktabs}

\title{Context-Aware Knowledge Distillation with Adaptive Weighting for Image Classification}
\author{Zhengda Li \\
School of Computing \\
Zhejiang University \\
HangZhou, China \\
\texttt{3220102159@zju.edu.cn}}
\date{August 30, 2025}

\begin{document}

\maketitle
\begin{abstract}
Knowledge distillation (KD) is a widely used technique to transfer knowledge from a large teacher network to
a smaller student model. Traditional KD uses a fixed balancing factor $\alpha$ as a hyperparameter to combine the hard-label cross-entropy
loss with the soft-label distillation loss. However, a static $\alpha$ is suboptimal because the optimal trade-off between
hard and soft supervision can vary during training. 

In this work, we propose an Adaptive Knowledge Distillation (AKD)
framework. First we try to make $\alpha$ as learnable parameter that can be automatically learned and optimized during training. Then we introduce a formula to reflect the gap between the student and the teacher to compute $\alpha$ dynamically, guided by student-teacher discrepancies, and further introduce a
Context-Aware Module (CAM) using MLP + Attention to adaptively reweight class-wise teacher outputs. Experiments
on CIFAR-10 with ResNet-50 as teacher and ResNet-18 as student demonstrate that our approach achieves superior
accuracy compared to fixed-weight KD baselines, and yields more stable convergence.
\end{abstract}

\section{Introduction}
Deep neural networks have achieved remarkable performance across computer vision tasks. However, large models
such as ResNet-50 are computationally expensive, limiting deployment on resource-constrained devices. Knowledge
distillation (KD) \cite{hinton2015distilling} addresses this by transferring knowledge from a large teacher model to a
smaller student network. The idea of KD involves transferring the valuable "knowledge" from a large, high-performing "teacher" model to a smaller, more efficient "student" network. Conventional KD optimizes a weighted sum of hard-label cross-entropy (CE) and soft-label
Kullback-Leibler divergence (KL) losses, with a fixed balancing coefficient $\alpha$. However, this fixed trade-off cannot
adapt to the evolving student-teacher relationship during training. For example, early in training, a student model has limited foundational knowledge and primarily needs hard supervision to learn fundamental concepts. Conversely, as the student becomes more proficient, it has already learned from hard labels. At this point, it benefits more from soft supervision to fine-tune its decision boundaries and capture complex inter-class relationships. If $\alpha$ remains too low, the student cannot fully leverage the rich knowledge from the teacher, which limits its potential for further performance improvement. 

In this paper, we propose an Adaptive Knowledge Distillation (AKD) framework with three major contributions:
\begin{itemize}
    \item Make $\alpha$ as a learnable parameter that balances CE and KD losses based on training context.
    \item Compute $\alpha$ dynamically by the gap between student and teacher model.
    \item A Context-Aware Module (CAM) that leverages student-teacher discrepancy and attention to generate
    class-wise adaptive weights for soft labels.
    \item Empirical evaluation on CIFAR-10 showing that AKD improves student performance over traditional KD.
\end{itemize}

\section{Methodology}
In this section, we provide a description of the proposed Adaptive Knowledge Distillation (AKD) framework, which combines dynamic and learnable weighting strategies to balance the contributions of hard-label and soft-label losses. We also introduce a novel Context-Aware Module (CAM) to enhance the teacher's supervision by focusing the student's attention on more informative classes.

\subsection{Knowledge Distillation Formulation}
KD using a weighted combination of the traditional hard-label cross-entropy (CE) loss and a soft-label loss, often based on Kullback-Leibler (KL) divergence. The standard KD loss is formulated as:
\begin{equation}
L = \alpha \cdot L_{\text{CE}}(y, p_s) + (1 - \alpha) \cdot T^2 \cdot KL(p_t^T || p_s^T),
\end{equation}
where:
\begin{itemize}
    \item $y$: The ground-truth label,
    \item $p_s$ and $p_t$: The student and teacher model outputs (i.e., class probabilities),
    \item $T$: The temperature used for smoothing the soft-labels,
    \item $\alpha$: A hyperparameter controlling the trade-off between the two loss terms.
\end{itemize}

The term $L_{\text{CE}}$ represents the typical cross-entropy loss between the student's predictions and the true labels, while the second term is the KL divergence between the teacher's softened outputs and the student's predictions. In traditional KD, $\alpha$ is typically a fixed constant, which balances the contribution from both losses. However, this static choice of $\alpha$ is suboptimal because the optimal balance between the hard and soft supervision changes over the course of training. Early in training, hard-label supervision may dominate as the student is far from the teacher’s performance, while later in training, the soft-label supervision from the teacher may provide more meaningful guidance.

\subsection{Learnable and Dynamic $\alpha$}
To address the issue of a static balancing factor, we propose two strategies for adaptively adjusting $\alpha$ during training.

\begin{itemize}
    \item \textbf{Learnable $\alpha$}: In this strategy, $\alpha$ is treated as a learnable parameter, optimized during training through gradient descent. Specifically, we parameterize $\alpha$ using a sigmoid function to constrain it to the range $(0, 1)$. This approach allows $\alpha$ to evolve dynamically based on the gradients from both loss terms, enabling the model to fine-tune the trade-off between hard and soft supervision. 
    \begin{equation}
    \alpha = \sigma(W_\alpha x + b_\alpha),
    \end{equation}
    where $W_\alpha$ and $b_\alpha$ are learned parameters, and $x$ is a feature vector that may depend on both the teacher and student outputs.
    
    \item \textbf{Dynamic $\alpha$}: The dynamic strategy involves adjusting $\alpha$ as a function of the discrepancy between the student’s and teacher’s predictions. Specifically, we define:
    \begin{equation}
    \alpha = \sigma(-k \cdot \| p_s - p_t \|),
    \end{equation}
    where $k$ is a hyperparameter that controls the steepness of the adjustment, and $\| p_s - p_t \|$ measures the discrepancy between the student and teacher's predictions. This approach emphasizes adjusting $\alpha$ based on how far the student is from the teacher, with the idea that a larger discrepancy should lead to a greater reliance on hard-label supervision. However, one challenge with this approach is the large difference in the magnitudes of the cross-entropy (CE) loss and the knowledge distillation (KD) loss, which may result in $\alpha$ quickly diminishing to zero.

    To address this issue, we substitute logits difference with probability distributions after applying softmax. The logits can have very large values, which, when squared, can produce a large discrepancy. Instead, we compute the difference between the student and teacher probabilities, which are constrained within the range $[0, 1]$. This ensures that the distance measure remains stable and in a reasonable range. Specifically, the discrepancy is computed as:
    \begin{equation}
    \text{dist} = \text{mean} \left( (p_s - p_t)^2 \right),
    \end{equation}
    where $p_s$ and $p_t$ are the softmax probabilities for the student and teacher, respectively. Since the probability values are within $[0, 1]$, the resulting distance measure also stays within this range, preventing $\alpha$ from diminishing too quickly.

    The $\alpha$ adjustment is then performed using the sigmoid function:
    \begin{equation}
    \alpha = \sigma(-k \cdot \text{dist}),
    \end{equation}
    where the value of $k$ determines the steepness of the sigmoid curve, controlling how aggressively $\alpha$ responds to the discrepancy between the teacher and student.

\end{itemize}

\subsection{Context-Aware Module (CAM)}
While the learnable and dynamic strategies for $\alpha$ help balance the losses, they do not directly account for the varying importance of different classes in the teacher’s supervision. To address this, we introduce a Context-Aware Module (CAM) that uses a class-wise attention mechanism to selectively focus the student on the most informative classes, enhancing the knowledge transfer process.

The CAM operates by first computing an attention vector $a$, which is generated using a multi-layer perceptron (MLP) that takes as input both the student’s and teacher’s logits, along with the discrepancy between the two. The attention vector $a$ is then applied to reweight the teacher's output probabilities. Specifically, the attention vector is computed as:
\begin{equation}
a = \sigma(\text{MLP}(p_s, p_t, |p_s - p_t|)),
\end{equation}
where $p_s$ and $p_t$ are the student’s and teacher’s predictions (logits), and $\sigma$ denotes the sigmoid activation function. The absolute difference $|p_s - p_t|$ captures the discrepancy between the teacher and student, which is used to weight the importance of each class during the knowledge transfer.

Once the attention vector $a$ is computed, it is applied to the teacher’s output probabilities $p_t$ to generate the context-aware teacher distribution:
\begin{equation}
p_t^{\text{CAM}} = \frac{a \odot p_t}{\sum_j (a \odot p_t)_j}.
\end{equation}
Here, $\odot$ denotes element-wise multiplication, and the denominator normalizes the attention-modified teacher distribution to ensure that it sums to 1. This operation selectively reweights the teacher’s outputs, prioritizing the classes where the teacher’s supervision is most beneficial for the student. The context-aware distribution encourages the student to focus more on the classes that are more informative or where the teacher’s guidance is more reliable.

This class-wise attention mechanism plays a critical role in guiding the student model’s learning process by ensuring that the student pays more attention to the regions of the feature space where the teacher’s output is more informative, thereby improving the overall performance of the student model.

\section{Experiments}
\subsection{Experimental Setup}
We conduct experiments on CIFAR-10 using ResNet-50 as the teacher and ResNet-18 as the student. All models are implemented in MindSpore architecture and executed on the Ascend platform by Huawei. The teacher model is pretrained on ImageNet and fine-tuned on CIFAR-10, while the student models are trained using SGD/Adam optimizers for 20 epochs.

We compare:
\begin{itemize}
    \item \textbf{KD (fixed $\alpha$)}: Traditional KD with $\alpha = 0.5$.
    \item \textbf{AKD (learnable $\alpha$)}: Joint optimization of $\alpha$.
    \item \textbf{AKD (dynamic $\alpha$)}: Distance-based adaptive $\alpha$.
    \item \textbf{AKD + CAM}: Proposed full model.
\end{itemize}

\subsection{Results}

\begin{table}[h]
\centering
\caption{Comparison of student performance under different KD strategies on CIFAR-10.}
\begin{tabular}{l c}
\toprule
Method & Accuracy (\%) \\
\midrule
KD (fixed $\alpha=0.5$) & 89.7 \\
AKD (learnable $\alpha$) & 90.4 \\
AKD (dynamic $\alpha$) & 90.7 \\
AKD + CAM (ours) & \textbf{91.5} \\
\bottomrule
\end{tabular}
\end{table}

\subsection{Experimental Results and Analysis}
\label{sec:results}

Our experimental results, summarized in Table 1. A baseline KD model with a fixed $\alpha$ of 0.5 achieved a validation accuracy of 89.7\%. By simply making $\alpha$ a learnable parameter, the student model's performance improved to 90.4\%. This shows that allowing the model to find the optimal trade-off between hard and soft supervision is beneficial. Furthermore, our approach of dynamically computing $\alpha$ based on the student-teacher discrepancy yielded an even higher accuracy of 90.7\%. Our full AKD framework incorporating the Context-Aware Module (CAM) reaches a final accuracy of \textbf{91.5\%}. This validates our hypothesis a more class-wise adaptive weighting, beyond a single global $\alpha$, is useful for effective knowledge transfer.

Figure 1 provides a detailed look into the training dynamics of our AKD + CAM model.
\begin{itemize}
    \item Loss Curves (Figure 1a): Both the training and validation loss consistently decrease over epochs, indicating that the model is learning effectively and generalizing well to the test data without significant signs of overfitting.
    \item Validation Accuracy (Figure 1b): The validation accuracy shows a steady and continuous increase throughout the training process, converging at a high performance level. This further confirms that our adaptive approach leads to a robust and stable training process.
    \item Adaptive $\alpha$ Evolution (Figure 1c): We can see that the average alpha value dynamically adjusts over the epochs. The curve shows a trend of how our framework balances hard and soft supervision over time. The specific shape of the curve reflects the CAM's intelligent adaptation to the student's learning stage, providing more guidance when needed and allowing for more independent learning as the student becomes more proficient.
\end{itemize}

\begin{figure}[h]
    \centering
    \includegraphics[width=0.95\linewidth]{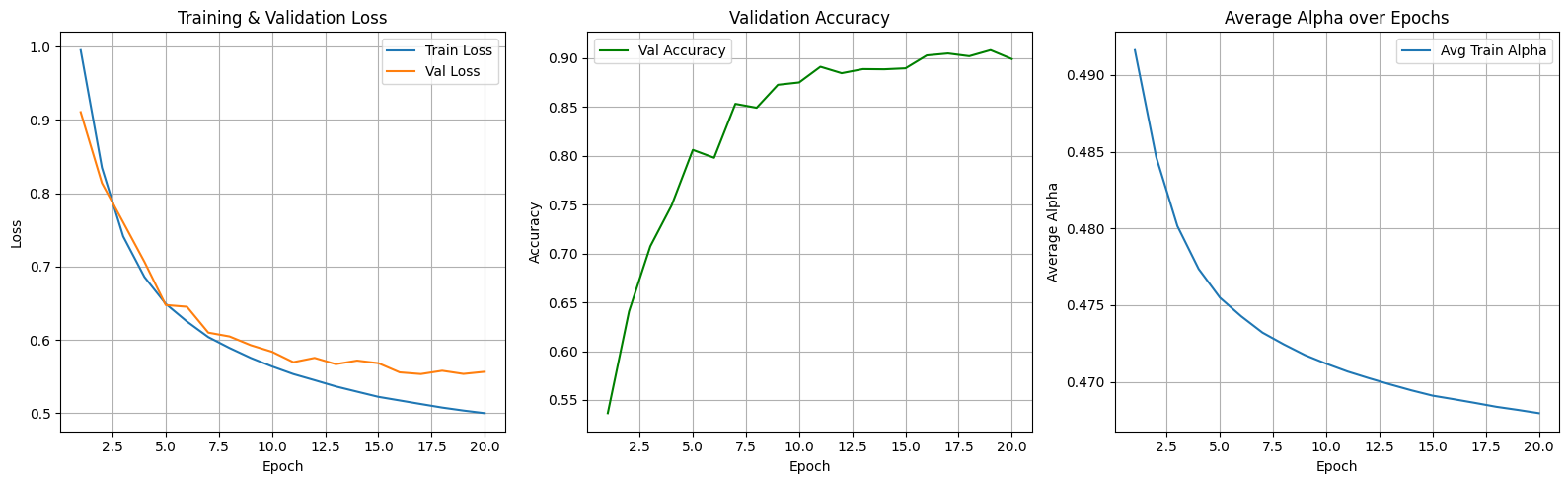}
    \caption{Training curves: (a) Loss, (b) Validation Accuracy, (c) Adaptive $\alpha$ evolution.}
\end{figure}

\subsection{Further Work}
In future work, we plan to explore the following directions:
\begin{itemize}
    \item \textbf{Extending to Other Models:} We will investigate the performance of our proposed AKD + CAM framework on other state-of-the-art architectures such as EfficientNet, VGG, and Transformer-based models. This will help assess the generalizability of the method across different model types.
    \item \textbf{Evaluation on Larger Datasets:} While our experiments have been conducted on CIFAR-10, we aim to evaluate our approach on larger and more complex datasets, such as ImageNet, to examine its scalability and effectiveness in real-world scenarios.
    \item \textbf{Combining with Other Regularization Techniques:} We will explore integrating our adaptive knowledge distillation with regularization methods like Dropout, Batch Normalization, or mixup, to further enhance the student model's robustness.
    \item \textbf{Exploring Teacher-Student Discrepancies:} Investigating more sophisticated ways to compute the discrepancy between teacher and student models, such as using more granular metrics or exploring layer-specific supervision, could further improve the knowledge transfer process.
\end{itemize}

\section{Conclusion}
We proposed an Adaptive Knowledge Distillation (AKD) framework with dynamic and learnable $\alpha$, and a novel
Context-Aware Module (CAM) to enhance teacher supervision. Experiments on CIFAR-10 demonstrate improved student
accuracy and stable convergence. Future work includes extending CAM to larger-scale datasets and multi-teacher
distillation.

\section{Acknowledgments}
Thanks for the support provided by MindSpore Community. All experiments proposed in this paper are implemented
based on the mindspore framework.

\end{document}